\pdfoutput=1

\documentclass[11pt]{article}

\usepackage{ACL2023}

\usepackage{times}
\usepackage{latexsym}

\usepackage[T1]{fontenc}

\usepackage[utf8]{inputenc}

\usepackage{microtype}

\usepackage{inconsolata}

\usepackage{multirow}
\usepackage{booktabs}
\usepackage{graphicx}

\title{With Good MT There is No Need For End-to-End:\\
A Case for Translate-then-Summarize Cross-lingual Summarization}

\author{Daniel Varab \\
  Novo Nordisk \\
  IT University of Copenhagen \\
  \texttt{djam@itu.dk} \\\And
  Christian Hardmeier \\
  IT University of Copenhagen \\
  \texttt{chrha@itu.dk} \\}

\begin{document}
\maketitle
\begin{abstract}
Recent work has suggested that end-to-end system designs for cross-lingual summarization are competitive solutions that perform on par or even better than traditional pipelined designs. A closer look at the evidence reveals that this intuition is based on the results of only a handful of languages or using underpowered pipeline baselines. In this work, we compare these two paradigms for cross-lingual summarization on 39 source languages into English and show that a simple \textit{translate-then-summarize} pipeline design consistently outperforms even an end-to-end system with access to enormous amounts of parallel data. For languages where our pipeline model does not perform well, we show that system performance is highly correlated with publicly distributed BLEU scores, allowing practitioners to establish the feasibility of a language pair a priori. Contrary to recent publication trends, our result suggests that the combination of individual progress of monolingual summarization and translation tasks offers better performance than an end-to-end system, suggesting that end-to-end designs should be considered with care. 
\end{abstract}

\section{Introduction}
Cross-lingual summarization (CLS) is the task of producing a summary of a text document that differs from the language it was written in, e.g. summarizing Turkish news or Danish product reviews in Hindi or English. This not only allows users fast access to information but also grants individuals access to information that is otherwise inaccessible. CLS is a challenging task as it must solve the challenges of both machine translation (MT) and summarization. There have historically been two approaches to the task;
\begin{itemize}
    \item Pipeline designs (translate, summarize)
    \item End-to-end designs (sequence-to-sequence)
\end{itemize}

\begin{figure}
    \centering
    \includegraphics[width=\columnwidth]{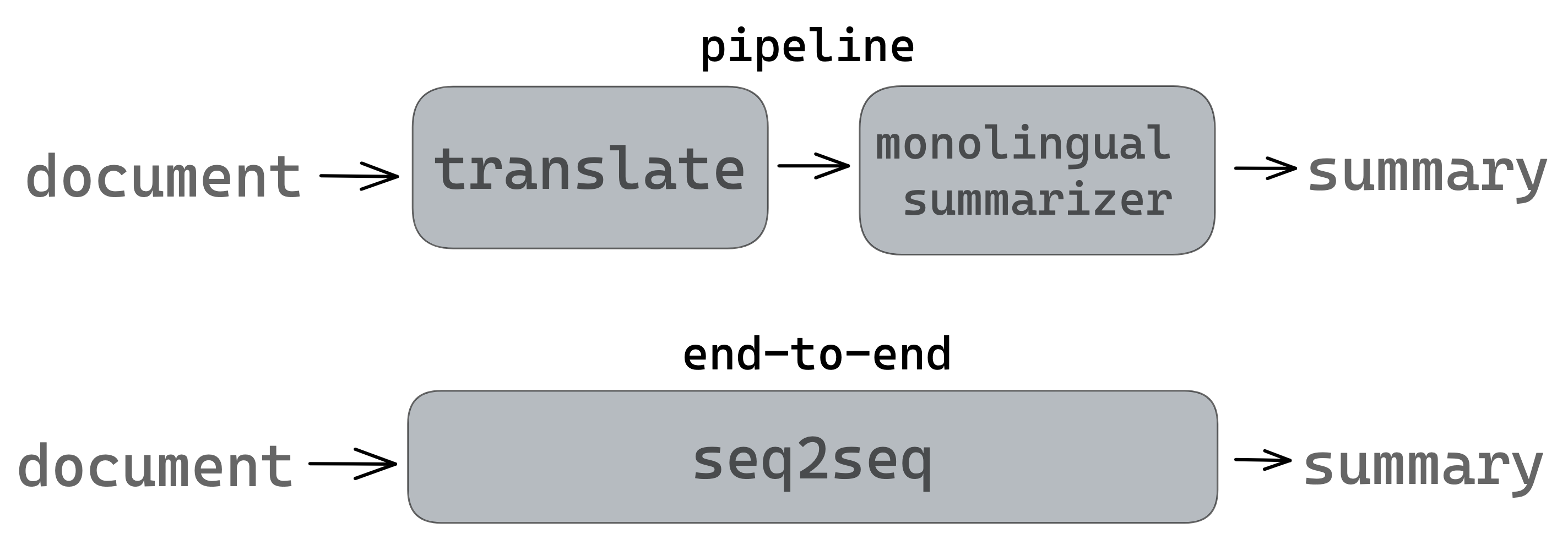}
    \caption{Pipeline versus end-to-end cross-lingual summarization designs. Pipeline-based systems perform cross-lingual summarization over two steps, first translating and then summarizing (or vice versa). End-to-end systems conflate translation and summarization by training a sequence-to-sequence to perform both tasks simultaneously.}
    \label{fig:my_label}
\end{figure}

Pipeline-based systems decompose CLS into two explicit steps, \textit{translation} and \textit{summarization}. This removes the necessity for parallel training data and enables taking advantage of ongoing innovations in translation and monolingual summarization research. The downside is the inherent effects of error propagation, where fx. a poor translation is forwarded to the subsequent summarization system, ultimately producing a bad summary. To circumvent this sequence-to-sequence designs have been proposed to avoid explicit translation and summarization steps altogether. With access to sufficiently large amounts of cross-lingual data, an end-to-end model can be trained to directly map an input document in one language, to a summary in another. The downside, however, is the sizable lack of CLS data, which does not occur naturally as opposed to the data of the implicit tasks: machine translation \citep{banon-etal-2020-paracrawl, aulamo-tiedemann-2019-opus, Fan2021BeyondEM} and monolingual summarization \citep{hermann-etal-cnndm, narayan-etal-2018-dont, grusky-etal-2018-newsroom, varab-schluter-2021-massivesumm, hasan-etal-2021-xl, scialom-etal-2020-mlsum}. In spite of this, a growing body of research is pushing the envelope on end-to-end CLS systems. \citep{zhu-etal-2019-ncls} and \citep{cao-etal-2020-jointly} created large synthetic CLS datasets using back-translation for English and Chinese. \citep{duan-etal-2019-zero} proposed directly distilling a system from existing monolingual summarization and translation systems using teacher forcing. The latest efforts have been put into collecting CLS data from online websites \citep{ladhak-etal-2020-wikilingua, perez-beltrachini-lapata-2021-models, DBLP:journals/corr/abs-2112-08804}.

\paragraph{Contributions} This paper investigates the immediate behaviors of two CLS paradigms on a wide range of languages and contributes with the following insights:

\begin{itemize}
    \item End-to-end systems do not convincingly outperform simple pipeline systems (translate-then-summarize) - even if provided with large amounts of data.
    \item Provided with a competitive MT system, pipeline systems outperform strong end-to-end systems by a large margin.
    \item Publicly distributed BLEU scores are reasonably correlated with pipeline performance and can be used to estimate the efficacy of a language pair for CLS a priori.
\end{itemize}

\section{Experiment}
We wish to evaluate a paradigm's ability to perform CLS and to produce evidence that helps resolves the status quo. Let $D_{s}=[w_1, \dots, w_n]$ be a text document consisting of words written in a source language $s$. The goal of a considered system is to produce a candidate summary $S_{t}$ written in a target language $t$, such that $S_{t}$ adequately summarizes the central information conveyed in $D_{s}$. In our experiments, we explore 39 different languages for $s$ but fixate $t$ = \textit{English}. We run two recently proposed designs for end-to-end (E2E) CLS and compare them to two simple but performant pipeline systems. We choose \textit{translate-then-summarize} (TTS) over \textit{summarize-then-translate} (STT) because STT requires monolingual summarization systems for each language, while translation systems are available for most language pairs. Using TTS, therefore, allows us to investigate more languages while taking advantage of progress in monolingual summarization research, which is primarily developed for English. We also argue that English is a suitable target language as it aligns well well with the practical goals of cross-lingual summarization: knowledge sharing through trade and international languages \citep{guerard1922short}.

\section{Models}

\subsection{Pipeline Systems}
Having chosen TTS it is sufficient to find a single summarization system. Since the summarization system will be compared against a sequence-to-sequence model we choose an abstractive summarization which also builds on a sequence-to-sequence architecture. We choose the BRIO \citet{liu-etal-2022-brio} system as it has recently shown strong performance across several standardized summarization benchmark datasets. For translation, we consider two systems. First, we consider the OPUS-MT models \citep{TiedemannThottingal:EAMT2020, junczys-dowmunt-etal-2018-marian}. OPUS-MT models are trained on the OPUS corpus \citep{aulamo-tiedemann-2019-opus} and support 180+ languages. Secondly, to explore the difference if using a more powerful MT system we consider the 418M parameter M2M100 \citep{Fan2021BeyondEM} model. This is a performant multilingual MT system that supports translation in any direction for 100 languages. We name these considered pipeline systems as follows:

\paragraph{TTS-weak} combines the OPUS-MT translation system with the abstractive summarization system BRIO. This system intends to investigate the effects of a lightweight MT system and quantify the effects of poor translations, and the performance drops resulting from cascading errors.

\paragraph{TTS-strong} combines the M2M100 translation system with the abstractive summarization system BRIO. This system acts as the competing alternative to an E2E system design. Results based on this system are the ones that will be considered when comparing the pipeline performance with E2E performance.


\subsection{End-to-End}
For end-to-end systems, consider the model proposed along with the CrossSum dataset \citep{DBLP:journals/corr/abs-2112-08804}. This model proposes fine-tuning over multiple language simultaneously using a multistage sampling technique to account for imbalance across languages. They report that training on multiple languages improves the performance of the system as a result of knowledge sharing between related languages. We also consider a zero-shot cross-lingual model recently proposed by \citet{perez-beltrachini-lapata-2021-models}. This model is trained using monolingual English data but freezes the embeddings and relies on the model to knowledge transfer to unseen languages.  We adopt the described training scheme but refrain from incorporating the meta-learning loss as the authors only reported minor improvements compared to not using it. We name the considered E2E systems:

\paragraph{E2E-ZS} is the latter zero-shot model proposed by \citet{perez-beltrachini-lapata-2021-models}. As text generation models are not known to transfer well to zero-shot settings, this system acts as a means to identify languages that are easy to transfer.

\paragraph{E2E-FT} is the former fine-tuned model proposed by \citet{DBLP:journals/corr/abs-2112-08804}. This is a strong model with access to large amounts of data in multiple languages during training and, therefore, acts as an E2E system for CLS. 

\section{Dataset}
We evaluate all systems on 39 languages in the validation set of CrossSum \citep{DBLP:journals/corr/abs-2112-08804}, a large-scale cross-lingual summarization dataset containing news articles from the multilingual British news outlet BBC. CrossSum consists of 1.7 million document-summary pairs and more than 1500+ language pairs. The corpus is built on top of XL–Sum \citep{hasan-etal-2021-xl}, a multilingual extension to XSum \citep{narayan-etal-2018-dont}, and is created by aligning articles written in different languages using the multilingual sentence embeddings \citep{feng-etal-2022-language}. CrossSumm contains summaries that like XL–Sum and XSum are short, often no longer than a single sentence. 

\begin{table}[h!]
  \small
  \centering
  \begin{tabular}{lrrrrr}
    \textbf{Language} & \multicolumn{4}{c|}{\textbf{ROUGE-1}} & \textbf{BLEU} \\
    \midrule
                & \shortstack{TTS\\\tiny{weak}} & \shortstack{TTS\\\tiny{strong}} & \shortstack{E2E\\\tiny{ZS}}& \shortstack{E2E\\\tiny{FT}} &    {} \\

    \midrule
    Somali      &     - &  23.3 &     18.3 &     \textbf{32.5} &  97.6 \\
    Tamil       &     - &  22.6 &     24.9 &     \textbf{30.7} &  89.1 \\
    \midrule
    Ukrainian   &  38.1 &  \textbf{39.0} &     25.7 &     33.5 &  64.1 \\
    Turkish     &  \textbf{42.2} &  41.4 &     29.8 &     34.9 &  63.5 \\
    Russian     &  39.6 &  \textbf{40.1} &     30.1 &     33.7 &  61.1 \\
    \midrule
    French      &  39.2 &  \textbf{39.3} &     29.7 &     33.2 &  57.5 \\
    Sinhala     &     - &  \textbf{33.4} &     17.7 &     30.4 &  51.2 \\
    \midrule
    Arabic      &  38.2 &  \textbf{38.5 }&     23.1 &     32.4 &  49.4 \\
    Bengali     &  27.1 &  25.3 &     14.2 &     \textbf{29.4} &  49.2 \\
    Marathi     &  13.6 &  \textbf{31.8} &     16.0 &     29.1 &  47.8 \\
    Indonesian  &  \textbf{42.0} &  41.8 &     28.9 &     35.5 &  47.7 \\
    Telugu      &     - &     - &     14.2 &     \textbf{29.4} &  47.6 \\
    Thai        &  \textbf{32.7} &     - &     17.6 &     30.6 &  47.2 \\
    Portuguese  &     - &  \textbf{36.8} &     25.5 &     32.2 &  46.9 \\
    Spanish     &  34.9 &  \textbf{36.2} &     27.8 &     31.4 &  46.4 \\
    Nepali      &     - &  24.7 &     24.8 &     \textbf{32.2} &  42.8 \\
    Japanese    &  34.8 &  \textbf{39.0} &     30.1 &     35.3 &  41.7 \\
    Hindi       &  32.9 &  \textbf{39.5} &     26.4 &     32.4 &  40.4 \\
    \midrule
    Korean      &  31.9 &  \textbf{34.4} &     26.9 &     32.0 &  39.2 \\
    Igbo        &  22.4 &  26.7 &     15.9 &     \textbf{27.6} &  38.5 \\
    Yoruba      &  17.5 &  20.4 &     18.2 &     \textbf{39.2} &  36.3 \\
    Welsh       &  24.6 &  23.1 &     15.9 &     \textbf{31.6} &  36.2 \\
    Hausa       &  18.9 &  23.7 &     17.3 &     \textbf{32.2} &  35.7 \\
    Azerbaijani &  21.4 &  28.5 &     20.0 &     \textbf{32.6} &  30.4 \\
    \midrule
    Tigrinya    &  17.2 &     - &     10.5 &     \textbf{20.3} &  29.9 \\
    Panjabi     &  18.0 &  17.2 &     14.3 &     \textbf{27.7} &  29.3 \\
    Oromo       &  11.9 &     - &     10.7 &     \textbf{23.4} &  27.3 \\
    Amharic     &     - &  20.2 &     16.0 &     \textbf{30.1} &  23.5 \\
    \midrule
    Persian     &     - &  \textbf{37.5} &     25.4 &     32.8 &     - \\
    Scottish    &     - &  15.5 &     16.7 &     \textbf{35.2} &     - \\
    Gujarati    &     - &  11.9 &     13.9 &     \textbf{29.7} &     - \\
    Kirghiz     &     - &     - &     16.8 &     \textbf{34.8} &     - \\
    Burmese     &     - &  14.2 &     20.4 &     \textbf{33.9} &     - \\
    Pushto      &     - &  33.3 &     25.7 &     \textbf{33.7} &     - \\
    Rundi       &  29.0 &     - &     19.4 &     \textbf{35.4} &     - \\
    Swahili     &     - &  \textbf{38.3} &     18.8 &     35.0 &     - \\
    Urdu        &  18.0 &  21.6 &     17.1 &     \textbf{31.7} &     - \\
    Uzbek       &     - &  17.0 &     17.9 &     \textbf{31.1} &     - \\
    Vietnamese  &  38.2 &  \textbf{42.0} &     29.7 &     34.8 &     - \\
    \end{tabular}
    \caption{}
    \caption{ROUGE-1 and BLEU scores for all four models, across all 39 languages. E2E$_{ZS}$ denotes the E2E zero-shot system, E2E$_{FT}$ the fine-tuned E2E system, TTS$_{strong}$ the TTS system using the M2M100 translation system, and TTS$_{weak}$, the TTS system using the OPUS-MT translation systems.}
    \label{results:mega-table}
\end{table}

\section{Results}
In Table~\ref{results:mega-table} we report the results of our experiments. Each language is associated with an F-1 ROUGE-1 \citep{lin-2004-rouge} and a BLEU score. We compute ROUGE scores with \texttt{sacrerouge} \citep{deutsch-roth-2020-sacrerouge} using the default parameters\footnote{ROUGE-1.5.5.pl -c 95 -m -r 1000 -n 2 -a}. The columns reflect the four considered models. The first three rows show average scores across subsets of languages filtered with BLEU scores. The rows provide detailed scores for each model on each language subset.  ROUGE scores that are empty are due to the language not being supported, while empty BLEU scores are simply unavailable. We do include results whenever possible for completeness.

\paragraph{Translation System Quality}
An obvious limitation of two-step systems is that poor translation systems are bound to produce poor-quality summaries. To quantify this relationship we search for available BLEU test scores \citep{papineni-etal-2002-bleu} for translation-based systems for all investigated languages. We collect scores for the OPUS-MT systems, but could to our surprise only find scores on subsets of languages or aggregated scores over multiple languages for M2M100 and mBART50. For the lack of better, we report OPUS-MT BLEU test scores for each language and emphasize that conclusions based on these scores on other models should be taken with great care. We also acknowledge that BLEU is not comparable across datasets, however, we do argue that the scores may be used as an approximation for the quality of a translation system.

\section{Analysis}
\begin{figure}
  \centering
  \includegraphics[width=\columnwidth]{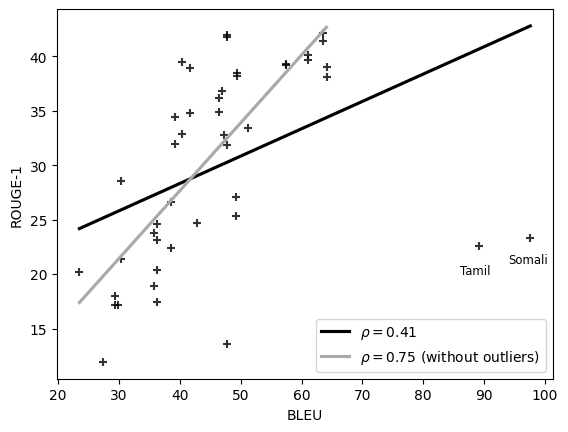}
  \caption{Collected BLEU scores on the x-axis and ROUGE-1 scores on the y-axis for TTS systems, including two outliers (Somali and Tamil) with suspiciously high BLEU scores. Removing the outliers further strengthens the relationship between the two metrics for TTS.}
  \label{results:tts-sbleu-vs-rouge-1}
\end{figure}

The results reveal three central insights. First, it is clear from the results of E2E-ZS that zero-shot is not feasible for CLS on the CrossSum dataset. Second, E2E-FT produces mostly low-to-mid scores with little low variance across languages. This model has the highest mean of 31.9.  Thirdly, TTS, despite having a slightly lower average of 28.5 and 29.6 between TTS-weak and TTS-strong respectively, these systems produce much higher scores on certain languages. A closer look reveals that despite E2E-FT scoring higher on average, both TTS systems frequently outperform E2Ef tune, and do so by a sizable margin. Conversely, when they do not they underperform significantly. Only four languages exhibit similar scores across the two paradigms, indicating a negative correlation between TTS-* and TT-FT. What we observe is that E2E-FT tune performs decently with little variation across languages, while TTS solutions either make or break it. Further inspection of the table suggests that the explanation for the TTS model’s performance can be explained by low-quality translations. In Figure~\ref{results:tts-sbleu-vs-rouge-1} we scatter plot translation and summarization scores for TTS systems and observe correlated behavior (Pearsons $\rho = 0.41$). A correlation that becomes visibly stronger if we allow removing suspicious BLEU scores (Somali and Tamil, $\rho = 0.75$).

\section{Conclusion}
In this paper, we question the recent trends in favor of end-to-end system design for CLS and address the current lack of fair comparisons to pipeline-based methods. We evaluate these two paradigms on many-to-one CLS from 39 source languages into English and show that despite the recent claims, and a general push toward end-to-end models, pipeline-based models remain a strong candidate for the task. We analyze the performance of pipeline-based models and show that performance is strongly correlated with translation quality (according to BLEU), and emphasize that this can be used to aid the decision-making for the development of real-world systems a priori using only public resources. With the results presented in this paper, we have produced evidence that allows practitioners and future researchers to re-consider the benefits of pipeline-based models.

\section{Limitations}
The experiments presented in this paper revolve around a single dataset of a specific summary type (single-sentence summaries). It is possible to imagine that if the experiments were run on another dataset the results would have produced other conclusions. However, due to the scarcity of cross-lingual summarization data and no other sizable datasets, it is not unclear how to broaden the experiment while still having enough data to support training a sequence-to-sequence model. We believe the empirical evidence presented in this paper adds valuable insights to peers and practitioners in the NLP community and that these results may serve as a counterweight to the focus on end-to-end system designs, highlighting an increasingly overlooked model option.

\clearpage
\bibliography{anthology,custom}

\begin{thebibliography}{27}
\expandafter\ifx\csname natexlab\endcsname\relax\def\natexlab#1{#1}\fi

\bibitem[{Aulamo and Tiedemann(2019)}]{aulamo-tiedemann-2019-opus}
Mikko Aulamo and J{\"o}rg Tiedemann. 2019.
\newblock \href {https://aclanthology.org/W19-6146} {The {OPUS} resource
  repository: An open package for creating parallel corpora and machine
  translation services}.
\newblock In \emph{Proceedings of the 22nd Nordic Conference on Computational
  Linguistics}, pages 389--394, Turku, Finland. Link{\"o}ping University
  Electronic Press.

\bibitem[{Ba{\~n}{\'o}n et~al.(2020)Ba{\~n}{\'o}n, Chen, Haddow, Heafield,
  Hoang, Espl{\`a}-Gomis, Forcada, Kamran, Kirefu, Koehn, Ortiz~Rojas,
  Pla~Sempere, Ram{\'\i}rez-S{\'a}nchez, Sarr{\'\i}as, Strelec, Thompson,
  Waites, Wiggins, and Zaragoza}]{banon-etal-2020-paracrawl}
Marta Ba{\~n}{\'o}n, Pinzhen Chen, Barry Haddow, Kenneth Heafield, Hieu Hoang,
  Miquel Espl{\`a}-Gomis, Mikel~L. Forcada, Amir Kamran, Faheem Kirefu, Philipp
  Koehn, Sergio Ortiz~Rojas, Leopoldo Pla~Sempere, Gema
  Ram{\'\i}rez-S{\'a}nchez, Elsa Sarr{\'\i}as, Marek Strelec, Brian Thompson,
  William Waites, Dion Wiggins, and Jaume Zaragoza. 2020.
\newblock \href {https://doi.org/10.18653/v1/2020.acl-main.417} {{P}ara{C}rawl:
  Web-scale acquisition of parallel corpora}.
\newblock In \emph{Proceedings of the 58th Annual Meeting of the Association
  for Computational Linguistics}, pages 4555--4567, Online. Association for
  Computational Linguistics.

\bibitem[{Bhattacharjee et~al.(2021)Bhattacharjee, Hasan, Ahmad, Li, Kang, and
  Shahriyar}]{DBLP:journals/corr/abs-2112-08804}
Abhik Bhattacharjee, Tahmid Hasan, Wasi~Uddin Ahmad, Yuan{-}Fang Li, Yong{-}Bin
  Kang, and Rifat Shahriyar. 2021.
\newblock \href {http://arxiv.org/abs/2112.08804} {Crosssum: Beyond
  english-centric cross-lingual abstractive text summarization for 1500+
  language pairs}.
\newblock \emph{CoRR}, abs/2112.08804.

\bibitem[{Cao et~al.(2020)Cao, Liu, and Wan}]{cao-etal-2020-jointly}
Yue Cao, Hui Liu, and Xiaojun Wan. 2020.
\newblock \href {https://doi.org/10.18653/v1/2020.acl-main.554} {Jointly
  learning to align and summarize for neural cross-lingual summarization}.
\newblock In \emph{Proceedings of the 58th Annual Meeting of the Association
  for Computational Linguistics}, pages 6220--6231, Online. Association for
  Computational Linguistics.

\bibitem[{Deutsch and Roth(2020)}]{deutsch-roth-2020-sacrerouge}
Daniel Deutsch and Dan Roth. 2020.
\newblock \href {https://doi.org/10.18653/v1/2020.nlposs-1.17} {{S}acre{ROUGE}:
  An open-source library for using and developing summarization evaluation
  metrics}.
\newblock In \emph{Proceedings of Second Workshop for NLP Open Source Software
  (NLP-OSS)}, pages 120--125, Online. Association for Computational
  Linguistics.

\bibitem[{Duan et~al.(2019)Duan, Yin, Zhang, Chen, and
  Luo}]{duan-etal-2019-zero}
Xiangyu Duan, Mingming Yin, Min Zhang, Boxing Chen, and Weihua Luo. 2019.
\newblock \href {https://doi.org/10.18653/v1/P19-1305} {Zero-shot cross-lingual
  abstractive sentence summarization through teaching generation and
  attention}.
\newblock In \emph{Proceedings of the 57th Annual Meeting of the Association
  for Computational Linguistics}, pages 3162--3172, Florence, Italy.
  Association for Computational Linguistics.

\bibitem[{Fan et~al.(2021)Fan, Bhosale, Schwenk, Ma, El-Kishky, Goyal, Baines,
  Çelebi, Wenzek, Chaudhary, Goyal, Birch, Liptchinsky, Edunov, Grave, Auli,
  and Joulin}]{Fan2021BeyondEM}
Angela Fan, Shruti Bhosale, Holger Schwenk, Zhiyi Ma, Ahmed El-Kishky,
  Siddharth Goyal, Mandeep Baines, Onur Çelebi, Guillaume Wenzek, Vishrav
  Chaudhary, Naman Goyal, Tom Birch, Vitaliy Liptchinsky, Sergey Edunov,
  Edouard Grave, Michael Auli, and Armand Joulin. 2021.
\newblock Beyond english-centric multilingual machine translation.
\newblock \emph{J. Mach. Learn. Res.}, 22:107:1--107:48.

\bibitem[{Feng et~al.(2022)Feng, Yang, Cer, Arivazhagan, and
  Wang}]{feng-etal-2022-language}
Fangxiaoyu Feng, Yinfei Yang, Daniel Cer, Naveen Arivazhagan, and Wei Wang.
  2022.
\newblock \href {https://doi.org/10.18653/v1/2022.acl-long.62}
  {Language-agnostic {BERT} sentence embedding}.
\newblock In \emph{Proceedings of the 60th Annual Meeting of the Association
  for Computational Linguistics (Volume 1: Long Papers)}, pages 878--891,
  Dublin, Ireland. Association for Computational Linguistics.

\bibitem[{Grusky et~al.(2018)Grusky, Naaman, and
  Artzi}]{grusky-etal-2018-newsroom}
Max Grusky, Mor Naaman, and Yoav Artzi. 2018.
\newblock \href {https://doi.org/10.18653/v1/N18-1065} {{N}ewsroom: A dataset
  of 1.3 million summaries with diverse extractive strategies}.
\newblock In \emph{Proceedings of the 2018 Conference of the North {A}merican
  Chapter of the Association for Computational Linguistics: Human Language
  Technologies, Volume 1 (Long Papers)}, pages 708--719, New Orleans,
  Louisiana. Association for Computational Linguistics.

\bibitem[{Gu{\'e}rard(1922)}]{guerard1922short}
Albert~L{\'e}on Gu{\'e}rard. 1922.
\newblock TF Unwin, Limited.

\bibitem[{Hasan et~al.(2021)Hasan, Bhattacharjee, Islam, Mubasshir, Li, Kang,
  Rahman, and Shahriyar}]{hasan-etal-2021-xl}
Tahmid Hasan, Abhik Bhattacharjee, Md.~Saiful Islam, Kazi Mubasshir, Yuan-Fang
  Li, Yong-Bin Kang, M.~Sohel Rahman, and Rifat Shahriyar. 2021.
\newblock \href {https://doi.org/10.18653/v1/2021.findings-acl.413} {{XL}-sum:
  Large-scale multilingual abstractive summarization for 44 languages}.
\newblock In \emph{Findings of the Association for Computational Linguistics:
  ACL-IJCNLP 2021}, pages 4693--4703, Online. Association for Computational
  Linguistics.

\bibitem[{Hermann et~al.(2015)Hermann, Kocisky, Grefenstette, Espeholt, Kay,
  Suleyman, and Blunsom}]{hermann-etal-cnndm}
Karl~Moritz Hermann, Tomas Kocisky, Edward Grefenstette, Lasse Espeholt, Will
  Kay, Mustafa Suleyman, and Phil Blunsom. 2015.
\newblock \href
  {https://proceedings.neurips.cc/paper/2015/file/afdec7005cc9f14302cd0474fd0f3c96-Paper.pdf}
  {Teaching machines to read and comprehend}.
\newblock In \emph{Advances in Neural Information Processing Systems},
  volume~28. Curran Associates, Inc.

\bibitem[{Junczys-Dowmunt et~al.(2018)Junczys-Dowmunt, Grundkiewicz, Dwojak,
  Hoang, Heafield, Neckermann, Seide, Germann, Aji, Bogoychev, Martins, and
  Birch}]{junczys-dowmunt-etal-2018-marian}
Marcin Junczys-Dowmunt, Roman Grundkiewicz, Tomasz Dwojak, Hieu Hoang, Kenneth
  Heafield, Tom Neckermann, Frank Seide, Ulrich Germann, Alham~Fikri Aji,
  Nikolay Bogoychev, Andr{\'e} F.~T. Martins, and Alexandra Birch. 2018.
\newblock \href {https://doi.org/10.18653/v1/P18-4020} {{M}arian: Fast neural
  machine translation in {C}++}.
\newblock In \emph{Proceedings of {ACL} 2018, System Demonstrations}, pages
  116--121, Melbourne, Australia. Association for Computational Linguistics.

\bibitem[{Ladhak et~al.(2020)Ladhak, Durmus, Cardie, and
  McKeown}]{ladhak-etal-2020-wikilingua}
Faisal Ladhak, Esin Durmus, Claire Cardie, and Kathleen McKeown. 2020.
\newblock \href {https://doi.org/10.18653/v1/2020.findings-emnlp.360}
  {{W}iki{L}ingua: A new benchmark dataset for cross-lingual abstractive
  summarization}.
\newblock In \emph{Findings of the Association for Computational Linguistics:
  EMNLP 2020}, pages 4034--4048, Online. Association for Computational
  Linguistics.

\bibitem[{Lin(2004)}]{lin-2004-rouge}
Chin-Yew Lin. 2004.
\newblock \href {https://aclanthology.org/W04-1013} {{ROUGE}: A package for
  automatic evaluation of summaries}.
\newblock In \emph{Text Summarization Branches Out}, pages 74--81, Barcelona,
  Spain. Association for Computational Linguistics.

\bibitem[{Liu et~al.(2022)Liu, Liu, Radev, and Neubig}]{liu-etal-2022-brio}
Yixin Liu, Pengfei Liu, Dragomir Radev, and Graham Neubig. 2022.
\newblock \href {https://doi.org/10.18653/v1/2022.acl-long.207} {{BRIO}:
  Bringing order to abstractive summarization}.
\newblock In \emph{Proceedings of the 60th Annual Meeting of the Association
  for Computational Linguistics (Volume 1: Long Papers)}, pages 2890--2903,
  Dublin, Ireland. Association for Computational Linguistics.

\bibitem[{Narayan et~al.(2018)Narayan, Cohen, and
  Lapata}]{narayan-etal-2018-dont}
Shashi Narayan, Shay~B. Cohen, and Mirella Lapata. 2018.
\newblock \href {https://doi.org/10.18653/v1/D18-1206} {Don{'}t give me the
  details, just the summary! topic-aware convolutional neural networks for
  extreme summarization}.
\newblock In \emph{Proceedings of the 2018 Conference on Empirical Methods in
  Natural Language Processing}, pages 1797--1807, Brussels, Belgium.
  Association for Computational Linguistics.

\bibitem[{Papineni et~al.(2002)Papineni, Roukos, Ward, and
  Zhu}]{papineni-etal-2002-bleu}
Kishore Papineni, Salim Roukos, Todd Ward, and Wei-Jing Zhu. 2002.
\newblock \href {https://doi.org/10.3115/1073083.1073135} {{B}leu: a method for
  automatic evaluation of machine translation}.
\newblock In \emph{Proceedings of the 40th Annual Meeting of the Association
  for Computational Linguistics}, pages 311--318, Philadelphia, Pennsylvania,
  USA. Association for Computational Linguistics.

\bibitem[{Paszke et~al.(2019)Paszke, Gross, Massa, Lerer, Bradbury, Chanan,
  Killeen, Lin, Gimelshein, Antiga et~al.}]{paszke2019pytorch}
Adam Paszke, Sam Gross, Francisco Massa, Adam Lerer, James Bradbury, Gregory
  Chanan, Trevor Killeen, Zeming Lin, Natalia Gimelshein, Luca Antiga, et~al.
  2019.
\newblock Pytorch: An imperative style, high-performance deep learning library.
\newblock \emph{Advances in neural information processing systems}, 32.

\bibitem[{Perez-Beltrachini and
  Lapata(2021)}]{perez-beltrachini-lapata-2021-models}
Laura Perez-Beltrachini and Mirella Lapata. 2021.
\newblock \href {https://doi.org/10.18653/v1/2021.emnlp-main.742} {Models and
  datasets for cross-lingual summarisation}.
\newblock In \emph{Proceedings of the 2021 Conference on Empirical Methods in
  Natural Language Processing}, pages 9408--9423, Online and Punta Cana,
  Dominican Republic. Association for Computational Linguistics.

\bibitem[{Rasley et~al.(2020)Rasley, Rajbhandari, Ruwase, and
  He}]{rasley2020deepspeed}
Jeff Rasley, Samyam Rajbhandari, Olatunji Ruwase, and Yuxiong He. 2020.
\newblock Deepspeed: System optimizations enable training deep learning models
  with over 100 billion parameters.
\newblock In \emph{Proceedings of the 26th ACM SIGKDD International Conference
  on Knowledge Discovery \& Data Mining}, pages 3505--3506.

\bibitem[{Reimers(2021)}]{reimers2021easynmt}
Nils Reimers. 2021.
\newblock Easynmt-easy to use, state-of-the-art neural machine translation.

\bibitem[{Scialom et~al.(2020)Scialom, Dray, Lamprier, Piwowarski, and
  Staiano}]{scialom-etal-2020-mlsum}
Thomas Scialom, Paul-Alexis Dray, Sylvain Lamprier, Benjamin Piwowarski, and
  Jacopo Staiano. 2020.
\newblock \href {https://doi.org/10.18653/v1/2020.emnlp-main.647} {{MLSUM}: The
  multilingual summarization corpus}.
\newblock In \emph{Proceedings of the 2020 Conference on Empirical Methods in
  Natural Language Processing (EMNLP)}, pages 8051--8067, Online. Association
  for Computational Linguistics.

\bibitem[{Tiedemann and Thottingal(2020)}]{TiedemannThottingal:EAMT2020}
J{\"o}rg Tiedemann and Santhosh Thottingal. 2020.
\newblock {OPUS-MT} — {B}uilding open translation services for the {W}orld.
\newblock In \emph{Proceedings of the 22nd Annual Conferenec of the European
  Association for Machine Translation (EAMT)}, Lisbon, Portugal.

\bibitem[{Varab and Schluter(2021)}]{varab-schluter-2021-massivesumm}
Daniel Varab and Natalie Schluter. 2021.
\newblock \href {https://doi.org/10.18653/v1/2021.emnlp-main.797}
  {{M}assive{S}umm: a very large-scale, very multilingual, news summarisation
  dataset}.
\newblock In \emph{Proceedings of the 2021 Conference on Empirical Methods in
  Natural Language Processing}, pages 10150--10161, Online and Punta Cana,
  Dominican Republic. Association for Computational Linguistics.

\bibitem[{Wolf et~al.(2020)Wolf, Debut, Sanh, Chaumond, Delangue, Moi, Cistac,
  Rault, Louf, Funtowicz, Davison, Shleifer, von Platen, Ma, Jernite, Plu, Xu,
  Le~Scao, Gugger, Drame, Lhoest, and Rush}]{wolf-etal-2020-transformers}
Thomas Wolf, Lysandre Debut, Victor Sanh, Julien Chaumond, Clement Delangue,
  Anthony Moi, Pierric Cistac, Tim Rault, Remi Louf, Morgan Funtowicz, Joe
  Davison, Sam Shleifer, Patrick von Platen, Clara Ma, Yacine Jernite, Julien
  Plu, Canwen Xu, Teven Le~Scao, Sylvain Gugger, Mariama Drame, Quentin Lhoest,
  and Alexander Rush. 2020.
\newblock \href {https://doi.org/10.18653/v1/2020.emnlp-demos.6} {Transformers:
  State-of-the-art natural language processing}.
\newblock In \emph{Proceedings of the 2020 Conference on Empirical Methods in
  Natural Language Processing: System Demonstrations}, pages 38--45, Online.
  Association for Computational Linguistics.

\bibitem[{Zhu et~al.(2019)Zhu, Wang, Wang, Zhou, Zhang, Wang, and
  Zong}]{zhu-etal-2019-ncls}
Junnan Zhu, Qian Wang, Yining Wang, Yu~Zhou, Jiajun Zhang, Shaonan Wang, and
  Chengqing Zong. 2019.
\newblock \href {https://doi.org/10.18653/v1/D19-1302} {{NCLS}: Neural
  cross-lingual summarization}.
\newblock In \emph{Proceedings of the 2019 Conference on Empirical Methods in
  Natural Language Processing and the 9th International Joint Conference on
  Natural Language Processing (EMNLP-IJCNLP)}, pages 3054--3064, Hong Kong,
  China. Association for Computational Linguistics.

\end{thebibliography}
\bibliographystyle{acl_natbib}

\appendix

\section{Appendix}
\subsection{Experimental Details}
\label{sec:experimental-details}
\subsection*{Abstractive Inference}
All models considered in this paper involve one (E2E) or two generation steps (TTS) which involve a few choices and a set of hyperparameters. For translation we translate documents in their entirety, sentence-by-sentence using the library EasyNMT\footnote{\url{github.com/UKPLab/EasyNMT}} \citep{reimers2021easynmt} which conveniently wraps the translation models considered in this work. We faced some issues with sentence segmentation in a few languages but changed the library code to make it work. For all summarization systems (including E2E) we truncate input documents to 512 tokens for all languages, use a beam size of 2, sample no longer than 128 tokens, and employ trigram blocking. When required by the model we add a decoder start token for English.

\subsection*{Training of Zero-Shot Model}
To train the zero-shot model described in the model section we adopt the methodology proposed by \citet{perez-beltrachini-lapata-2021-models} and implement it using Huggingface's \texttt{transformers} \citep{wolf-etal-2020-transformers}, \texttt{DeepSpeed} \citep{rasley2020deepspeed}, and of course \texttt{PyTorch} \citep{paszke2019pytorch}. We freeze the embeddings of the encoder and decoder of \texttt{mBART50} but do not prune the vocabulary. We also do not apply the proposed meta-learning algorithm LF-MALM for the sake of simplicity. We train the model with cross-entropy for 50.000 steps with a batch size of 32 using fp16 mixed-precision training and evaluate and save the model every 1000 steps. We also run a linear learning rate scheduler with warmup for 5000 steps (5e-5). Results are produced using the model with the lowest loss (1.886). This model took approximately 3 days to run on two NVIDIA T4 Tensor Core GPUs using \texttt{DeepSpeed}.

\end{document}